\documentclass[letterpaper, 10pt, journal, twoside]{IEEEtran}

\IEEEoverridecommandlockouts % This command is only needed if
\makeatletter

\let\proof\@undefined
\let\endproof\@undefined
\makeatother

\usepackage{hyperref}
\hypersetup{
  colorlinks,
  citecolor=black,
  filecolor=black,
  linkcolor=black,
  urlcolor=black,
  pdfauthor={},
  pdfsubject={},
  pdftitle={}
}

\usepackage{graphicx}
\usepackage{setspace}
\usepackage{epstopdf}
\usepackage{float}
\usepackage{multirow,tabularx,makecell}
\usepackage{siunitx}
\usepackage{cellspace}
\newcolumntype{D}{>{\hfill}N{3}{2}<{\hfill}}

\usepackage{algorithm}
\usepackage{etoolbox}\AtBeginEnvironment{algorithmic}{\scriptsize}
\usepackage[noend]{algpseudocode}

\usepackage{wrapfig}
\usepackage[font={small}]{caption}
\usepackage{scalefnt}

\usepackage{subfig}
\usepackage[export]{adjustbox}
\usepackage{environ}
\usepackage[para, bottom]{footmisc}
\usepackage{color}
\usepackage{xcolor}
\usepackage{url}

\usepackage{amsmath}
\usepackage{amssymb,amsmath}
\usepackage{mathtools}
\usepackage{siunitx}
\usepackage[nameinlink, noabbrev, capitalize]{cleveref}
\usepackage{cite}

\usepackage{trimclip}

\usepackage{isotope}
\usepackage{listings}
\makeatletter
\def\lst@makecaption{%
  \def\@captype{table}%
  \@makecaption
}
\makeatother

\makeatletter
\def\BState{\State\hskip-\ALG@thistlm}
\makeatother

\definecolor{dark_green}{rgb}{0.0, 0.6, 0.0}

\usepackage{tikz}
\usepackage{pgfplots}
\usepgfplotslibrary{groupplots}
\usetikzlibrary{backgrounds,arrows,automata,shapes,positioning,calc,through}
\pgfplotsset{compat=newest}
\pgfdeclarelayer{background}
\pgfdeclarelayer{foreground}
\pgfsetlayers{background,main,foreground}

\tikzset{
  state/.style={
    rectangle,
    draw=black, very thick,
    minimum height=1.0em,
    text centered,
  },
  legend_box/.style={
    rectangle,
    draw=black,
    text centered,
  },
  normalstate/.style={
    rectangle,
    draw=black, very thick,
    minimum height=2.9em,
    minimum width=6.25em,
    text centered,
  },
  finalstate/.style={
    rectangle,
    double=white,
    double distance=0.1em,
    inner sep=0.2em,
    draw=black, very thick,
    minimum height=2.90em,
    minimum width=6.25em,
    text centered,
  },
  initialstate/.style={
    rectangle,
    double=white,
    double distance=0.1em,
    inner sep=0.2em,
    draw=black, very thick,
    minimum height=2.90em,
    minimum width=6.25em,
    text centered,
  },
  point/.style={
    circle,
    inner sep=0pt,
    minimum size=3pt,
    fill=red
  },
  adder/.style={
    circle,
    inner sep=2pt,
    minimum size=0.3in,
    draw=black, very thick,
    text centered
  },
  arrow/.style={
    thick,
    ->,
  >=stealth},
  darrow/.style={
    thick,
    <->,
  >=stealth},
  block/.style={
        draw,
        rectangle,
        rounded corners,
        inner sep=0pt,
        fill=white,
        fill opacity=1.0,
        text opacity=1.0
    }
}
\usepackage{soul}

\usepackage{scalerel}
\usetikzlibrary{svg.path}
\definecolor{orcidlogocol}{HTML}{A6CE39}
\tikzset{
  orcidlogo/.pic={
    \fill[orcidlogocol] svg{M256,128c0,70.7-57.3,128-128,128C57.3,256,0,198.7,0,128C0,57.3,57.3,0,128,0C198.7,0,256,57.3,256,128z};
    \fill[white] svg{M86.3,186.2H70.9V79.1h15.4v48.4V186.2z}
    svg{M108.9,79.1h41.6c39.6,0,57,28.3,57,53.6c0,27.5-21.5,53.6-56.8,53.6h-41.8V79.1z M124.3,172.4h24.5c34.9,0,42.9-26.5,42.9-39.7c0-21.5-13.7-39.7-43.7-39.7h-23.7V172.4z}
    svg{M88.7,56.8c0,5.5-4.5,10.1-10.1,10.1c-5.6,0-10.1-4.6-10.1-10.1c0-5.6,4.5-10.1,10.1-10.1C84.2,46.7,88.7,51.3,88.7,56.8z};
  }
}
\newcommand\orcidicon[1]{\href{https://orcid.org/#1}{\mbox{\scalerel*{
        \begin{tikzpicture}[yscale=-1,transform shape]
          \pic{orcidlogo};
        \end{tikzpicture}
}{|}}}}

\title{Autonomous Aerial Filming with Distributed 
\\ Lighting by a Team of Unmanned \\ Aerial Vehicles}

\author{V\'{i}t Kr\'{a}tk\'{y}$^{1\orcidicon{0000-0002-1914-742X}}$, Alfonso Alc\'{a}ntara$^{2\orcidicon{0000-0002-4307-0995}}$, Jes\'{u}s Capit\'{a}n$^{2\orcidicon{0000-0002-7534-0187}}$,
Petr \v{S}t\v{e}p\'{a}n$^{1\orcidicon{0000-0002-7444-3264}}$,
Martin Saska$^{1\orcidicon{0000-0001-7106-3816}}$ and
An\'{i}bal Ollero$^{2\orcidicon{0000-0003-2155-2472}}$
  \thanks{Manuscript received: February 24, 2021; Revised May 26, 2021; Accepted July 3, 2021.}
  \thanks{
  This paper was recommended for publication by Editor M. Ani Hsieh upon evaluation of the Associate Editor and Reviewers' comments.
  This work was supported by EU project AERIAL-CORE (H2020-2019-871479), by MULTICOP (US-1265072) in FEDER-Junta de Andalucia Programme, by project no. DG18P02OVV069 in program NAKI II, by CTU grant no SGS20/174/OHK3/3T/13, and by OP VVV funded project CZ.02.1.01/0.0/0.0/16 019/0000765 "Research Center for Informatics". \textit{(V\'{i}t Kr\'{a}tk\'{y} and Alfonso Alc\'{a}ntara are co-first authors.)}}
  \thanks{$^1$V\'{i}t Kr\'{a}tk\'{y}, Petr \v{S}t\v{e}p\'{a}n, and Martin Saska are with Faculty of Electrical Engineering, Czech Technical University in Prague, Czech Republic, {\tt\footnotesize\{\href{mailto:vit.kratky@fel.cvut.cz}{kratkvit}|\href{mailto:stepan@fel.cvut.cz}{stepan}|\href{mailto:martin.saska@fel.cvut.cz}{martin.saska}\}@fel.cvut.cz}.}
  \thanks{$^2$Alfonso Alc\'{a}ntara, Jes\'{u}s Capit\'{a}n, and An\'{i}bal Ollero are with GRVC Robotics Laboratory, University of Seville, Spain {\tt\footnotesize\{\href{mailto:aamarin@us.es}{aamarin}|\href{mailto:jcapitan@us.es}{jcapitan}|\href{mailto:aollero@us.es}{aollero}\}@us.es}. 
  \newline 
  Digital Object Identifier (DOI): see top of this page.}
}

\begin{document}

% %%{ COPYRIGHT NOTICE
\newcommand{\PREPRINTYEAR}{2021}
\newcommand{\PREPRINTPUBLISHER}{IEEE}

\onecolumn
\pagenumbering{gobble}
{
  \topskip0pt
  \vspace*{\fill}
  \centering
  \LARGE{%
    \copyright{} \PREPRINTYEAR~\PREPRINTPUBLISHER\\\vspace{1cm}
	Personal use of this material is permitted.
	Permission from \PREPRINTPUBLISHER~must be obtained for all other uses, in any current or future media, including reprinting or republishing this material for advertising or promotional purposes, creating new collective works, for resale or redistribution to servers or lists, or reuse of any copyrighted component of this work in other works.}
	\vspace*{\fill}
}

\twocolumn % COMMENT FOR SINGLE-COLUMN ARTICLES
\pagenumbering{arabic}

\markboth{\copyright{} \PREPRINTPUBLISHER, \PREPRINTYEAR. Accepted to IEEE RA-L. DOI: \href{https://doi.org/10.1109/LRA.2021.3098811}{10.1109/LRA.2021.3098811}}{\copyright{} \PREPRINTPUBLISHER, \PREPRINTYEAR. Accepted to IEEE RA-L. DOI: \href{https://doi.org/10.1109/LRA.2021.3098811}{10.1109/LRA.2021.3098811}}
% %%}

\maketitle

%%{ ABSTRACT

\begin{abstract}
This paper describes a method for autonomous aerial cinematography with distributed lighting by a team of unmanned aerial vehicles (UAVs). 
Although camera-carrying multi-rotor helicopters have become commonplace in cinematography, their usage is limited to scenarios with sufficient natural light or of lighting provided by static artificial lights.
We propose to use a formation of unmanned aerial vehicles as a tool for filming a target under illumination from various directions, which is one of the fundamental techniques of traditional cinematography. 
We decompose the multi-UAV trajectory optimization problem to tackle non-linear cinematographic aspects and obstacle avoidance at separate stages, which allows us to re-plan in real time and react to changes in dynamic environments. 
The performance of our method has been evaluated in realistic simulation scenarios and field experiments, where we show how it increases the quality of the shots and that it is capable of planning safe trajectories even in cluttered environments.

\end{abstract}


%%{ MAIN TEXT

\section{INTRODUCTION}

\begin{figure}
    \centering
    \includegraphics[width=\columnwidth]{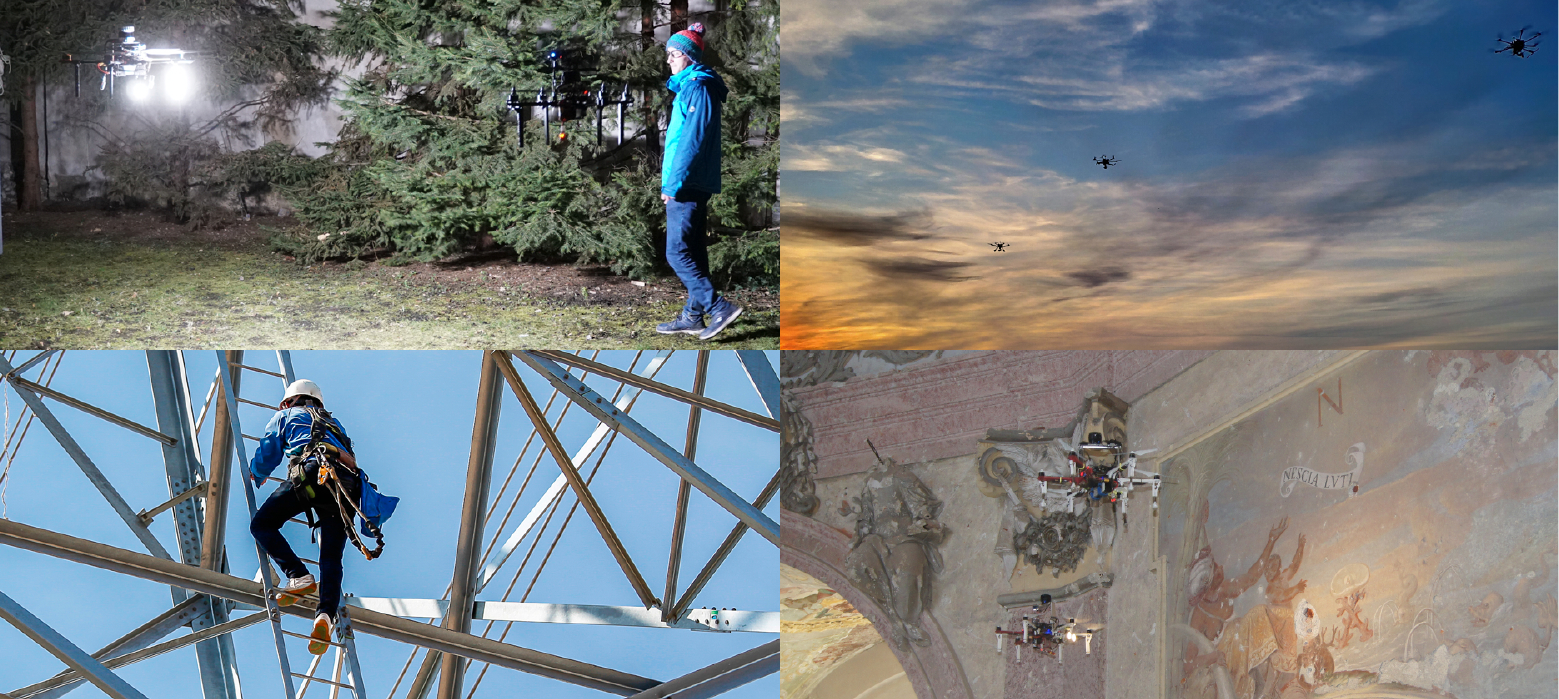}
    \caption{UAV filming applications to provide external lighting; to capture smooth shots outdoors; and to monitor dangerous maintenance operations at electrical lines\protect\footnotemark. Pictures were obtained within AERIAL-CORE and DRONUMENT projects, for which the proposed technology is being developed. Videos of the work in this paper can be seen on the multimedia material page (\url{http://mrs.felk.cvut.cz/papers/aerial-filming}).}
    \label{fig:uav_intro}
    \vspace{-0.5cm}
\end{figure}

\IEEEPARstart{T}{he} interest in \emph{Unmanned Aerial Vehicles} (UAVs) for aerial photography and filming is growing fast~\cite{mademlis_tb19,sabirova_20,Jeon_20, caraballo_iros20, moreno2020jibcrane}. This is mainly due to their manoeuvrability and the capacity to create unique shots when compared to standard cameras. The use of UAVs as flying cameras presents not only a remarkable potential for recreational cinematography, but also for the monitoring of inspection operations in outdoor infrastructures with complex access. For instance, the EU-funded project, AERIAL-CORE, proposes UAVs to surveil the safety of human workers during maintenance operations of electrical power lines (see \autoref{fig:uav_intro}). 
In this industrial setup, a high-quality video is key, as it is used by supervising ground operators to monitor safety during the maintenance work.
Multi-UAV teams expand upon these possibilities as they could provide alternative points of view or even supplementary illumination. 
Similarly in our DRONUMENT project of NAKI II program, efficient variable illumination plays a key role for documentation of historical buildings interiors.

Proper lighting techniques are fundamental in bringing out details in an image and in creating more natural-looking film scenes. 
Thus, cinematography sets are packed with different lighting sources, as digital sensors are not as reactive to light as the human eye. 
This can also be relevant in monitoring maintenance operations scheduled at times of the day with poor illumination. 
Although aerial cinematography has been attractive to the scientific community as of late, lighting techniques have yet to be applied to improve the performance of filming.
Filmmakers apply many types of lighting techniques making use of various equipment. 
In this work, we only consider direct lighting techniques that do not require additional equipment apart from light sources.
Although static lights could produce more pleasant footage in some situations, we believe that UAVs are not optimal for this purpose.     
Therefore, we only use UAVs as dynamic sources of light to provide lighting to a dynamic scene.

\footnotetext{https://aerial-core.eu}

% our method
In this context, navigating a team of UAVs for filming tasks with distributed lighting is complex. Smooth and safe trajectories are required to achieve pleasing shots that do not compromise safety in dynamic scenarios. We propose a method for online trajectory planning and execution with multiple UAVs. Our team obeys a leader-follower scheme where the formation leader carries an onboard camera to film a moving target and the followers generate trajectories that enable distributed lighting of the target, while maintaining desired lighting angles. We formulate a non-linear, optimization-based method that plans visually pleasant trajectories for the filming UAV and distributes the others in a specified formation. Moreover, we tackle safety by including a systematic framework for obstacle avoidance. Safe flight corridors for the UAVs are generated by forming sets of convex polyhedrons that model free space. Optimal and safe trajectories are thereafter computed within these convex sets. 

\subsection{Related works} 
% single UAV works in cinematography
There have been several works focusing on filming dynamic scenes with a single UAV. Commercial products exist (e.g., \textit{DJI Mavic}~\cite{mavic} or \textit{Skydio}~\cite{skydio}) that implement semi-autonomous functionalities, such as \textit{auto-follow} features for tracking an actor with collision avoidance. However, they do not address cinematographic principles. An integrated system for outdoor cinematography combining vision-based target localization with trajectory planning and collision avoidance has been proposed~\cite{Bonatti2019,bonatti_jfr20}. Smoothness is achieved by minimizing trajectory jerk; shot quality by defining objective curves that fulfill relative angles with respect to the actor. Optimal trajectories for cinematography have also been computed in real-time through receding horizon optimization with non-linear constraints~\cite{naegeli_ral17}. A user inputs framing objectives for the targets on the image to minimize errors on the image target projections, sizes, and relative viewing angles. Some authors have approached UAV cinematography by applying machine learning~\cite{passalis_18, dang_20}. Particularly, such techniques have been applied to demonstrations imitating professional cameraman's behaviors~\cite{huang_icra19} or for reinforcement learning to achieve visually pleasant shots~\cite{Gschwindt2019}. These works have presented valuable results for online trajectory planning, although they have not addressed the specific complexities for multi-UAV systems. 

% Multiples UAVs in cinematography
Regarding the methodology for multiple UAVs, a non-linear optimization problem was solved in receding horizon in~\cite{naegeli_tg17}, where collision avoidance to filmed actors and mutual collisions of UAVs were considered. Aesthetic objectives are introduced by the user as virtual reference trails. A specific camera parameter space is proposed in~\cite{galvane_tg18} to ensure cinematographic properties and to fulfill dynamic constraints along the trajectories. The motion of multiple UAVs around dynamic targets is coordinated through a centralized master-slave approach. A greedy framework for multi-UAV camera coordination is proposed in~\cite{Bucker2020}. A decentralized planner computes UAV trajectories considering smoothness, shot diversity, collision avoidance, and mutual visibility.  
% previous works in multi-UAV cinematography
We have also addressed the trajectory planning for multi-UAV cinematography in previous work. We presented an architecture to execute cinematographic shots (with different types of camera motion) using multiple UAVs~\cite{alcantara_access20} and developed a distributed method to plan optimal trajectories reducing jerky camera movements~\cite{alcantara2020optimal}. In this paper, our focus is on the specifics of outdoor and dynamic settings when compared to indoor scenarios~\cite{naegeli_tg17}. Therefore, we have integrated obstacle avoidance in a more fundamental manner using local maps. Moreover, a novel problem with respect to previous work has been introduced, as we perform scene illumination with multiple UAVs to increase the quality of image shots.  

% none of them with lighting
The modification of lighting angles to improve images is fundamental in cinematography~\cite{hall2015understanding}.
A camera with an onboard light on a UAV  can compensate for insufficient lighting, but positioning lights at different angles with respect to the camera axis would require several UAVs. 
Despite the unquestionable importance of lighting for shot quality, its usage for aerial cinematography is not well-studied.
Utilizing UAVs to provide supplementary illumination has been proposed for building documentation tasks~\cite{petracek2020ral} or tunnel inspection~\cite{petrlik2020ral}.
A formation with a filming UAV and  others carrying lights was deployed to document the overshadowed parts of historical buildings~\cite{saska2017etfa}.
A similar system has been used to carry out specialized documentation techniques~\cite{kratky2020ral}. However, these works have proposed lighting for tasks in static scenes, whereas the present paper deals with filming of moving targets in dynamic and potentially cluttered environments, e.g., to monitor inspection operations in large outdoor infrastructures.   

%methods that use similar obstacle avoidance approaches
In order to guarantee safe trajectories in multi-UAV cinematography, most works~\cite{naegeli_tg17,galvane_tg18,alcantara2020optimal} only consider collision avoidance with actors, other UAVs, or static objects that can be modelled with previously known no-fly zones. The work in~\cite{bonatti_jfr20} integrates local mapping with onboard sensors to penalize proximity to obstacles and solves an unconstrained optimization problem. Another approach to obstacle avoidance applied for standard UAV trajectory planning is to create a convex representation of free space via a set of linear inequality constraints~\cite{chen_icra16,Mohta2018,Liu2017,Tordesillas2019}, to obtain a QP formulation for real-time motion planning.
We have been inspired by these single-UAV works to develop a fundamental framework for the representation of obstacles in our non-linear optimization problem for multi-UAV cinematography. 

\subsection{Contributions}

Our main contributions are summarized as the following:
\begin{itemize}

    \item We formulate a novel optimization problem for aerial filming with distributed lighting. Using a leader-follower scheme, we plan and execute trajectories in a distributed manner. Optimization is run in receding horizon to compute smooth trajectories with pleasing footage for the UAV filming (the leader), which takes shots of a dynamic target indicated by an external user. The followers compute their trajectories to maintain a formation with specified lighting angles on the target. 
    
    \item We propose a new method to tackle non-convex trajectory optimization with obstacle avoidance in real time. We decompose the problem in two parts. Non-linear cinematographic aspects are formulated in a problem without obstacle avoidance to generate reference trajectories. These are used to generate collision-free regions which are convex and to transform the problem into a final QP optimization task. 
    
    \item We present experimental results for different types of cinematographic shots. We prove that our method is capable of computing smooth trajectories for reducing jerky movements and show that the distributed formation improves the illumination of footage. The system is evaluated with field experiments and also in various realistic simulated scenarios, including the filming of a moving target in a cluttered environment.
\end{itemize}

\section{SYSTEM OVERVIEW}
\label{sec:system_overview}

\begin{figure*}[!htb] 
  \centering
    % \begin{tikzpicture}[node distance=2cm, transform canvas={scale=0.9}]
  \definecolor{color_red}{HTML}{A30D00}
  \definecolor{color_green}{rgb}{0, .522, .243}
  \definecolor{color_blue}{rgb}{0, 0, .9}

\pgfdeclarelayer{background}
\pgfdeclarelayer{foreground}
\pgfsetlayers{background,main,foreground}
  \begin{tikzpicture}[node distance=2cm]

    \pgfmathsetmacro{\vshift}{0.07em} % shift in vertical direction
    \pgfmathsetmacro{\hshift}{0.11em} % shift in horizontal direction
    \pgfmathsetmacro{\vblocksize}{1.9} % height of block with space
    \pgfmathsetmacro{\hblocksize}{2.0} % height of block with space
    \pgfmathsetmacro{\harrowshiftem}{0.8em} % horizontal shift of arrows
    \pgfmathsetmacro{\varrowshiftem}{0.6em} % vertical shift of arrows
    \pgfmathsetmacro{\varrowshift}{0.01em} % vertical shift of arrows
    \pgfmathsetmacro{\vlabelshift}{0.01em} % vertical shift of arrows
    \pgfmathsetmacro{\hlabelshift}{0.025em} % vertical shift of arrows
    \pgfmathsetmacro{\harrowshift}{0.03em} % vertical shift of arrows
    \pgfmathsetmacro{\harrowbend}{0.38mm} % horizontal shift of arrows
    \pgfmathsetmacro{\varrowbend}{0.038em} % horizontal shift of arrows
    % \pgfmathsetmacro{\eventlabel}{} % event mark 

  \begin{pgfonlayer}{foreground}
    %%%{ Nodes definition

    \node[block, shift = {(0.0, 0.0)}] (human_director) {
      \begin{tabular}{c}
        \small Human \\
        \small director
      \end{tabular}
    };

    \node[block, right of=human_director, shift = {(\hshift+0.3, 0.0)}] (cinematographic_trajectory_generator) {
      \begin{tabular}{c}
        \small Cinematographic \\
        \small trajectory \\ 
        \small generator
      \end{tabular}
    };

    \node[block, below of=human_director, shift = {(0.0, 0.0)}] (target_detection) {
      \begin{tabular}{c}
        \small Target detection \\
        \small \& trajectory \\
        \small estimation
      \end{tabular}
    };

    \node[block, below of=cinematographic_trajectory_generator, shift = {(0.0, 0.0)}] (lighting_trajectory_generator) {
      \begin{tabular}{c}
        \small Lighting \\
        \small trajectory \\ 
        \small generator
      \end{tabular}
    };

    \node[block, right of=cinematographic_trajectory_generator, shift = {(0.9*\hshift, -1.0)}] (path_generator) {
      \begin{tabular}{c}
        \small Collision-free  \\
        \small path generator
      \end{tabular}
    };

    \node[block, right of=path_generator, shift = {(\hshift, 0.0)}] (safe_corridor_generator) {
      \begin{tabular}{c}
        \small Safe corridor \\
        \small generator
      \end{tabular}
    };

    \node[block, right of=safe_corridor_generator, shift = {(\hshift, 0.0)}] (trajectory_optimizer) {
      \begin{tabular}{c}
        \small Trajectory \\
        \small optimization
      \end{tabular}
    };

    \node[block, right of=trajectory_optimizer, shift = {(0.8*\hshift, 0.0)}] (trajectory_tracker) {
      \begin{tabular}{c}
        \small Trajectory \\
        \small tracker
      \end{tabular}
    };

    \node[above of=safe_corridor_generator, shift = {(-0.85*\hshift, -0.9)}] (map) {
      \begin{tabular}{c}
        \small Map of environment 
      \end{tabular}
    };

    \node[right of=map, shift = {(1.85*\hshift, 0.0)}] (dynamic_constraints) {
      \begin{tabular}{c}
        \small Dynamic constraints 
      \end{tabular}
    };
    %%%}

  %%%{ Arrows definition

    \draw [arrow, color=black] ($(human_director.east) - (0, -2*\varrowshift)$) -- ($(cinematographic_trajectory_generator.west) - (0, -2*\varrowshift)$) node[midway, above, shift={(0.0, 0.0*\vlabelshift)}] {\small $C_s$} ;
    \draw [arrow, color=black] ($(human_director.east) - (0, 2*\varrowshift)$) -| ($(human_director.east) + (1.0, -1.0)$) |- ($(lighting_trajectory_generator.west) - (0, -2*\varrowshift)$) node[midway, above, shift={(1.0*\hlabelshift, 4.8*\vlabelshift)}] {\small $C_l$ } ;
    \draw [arrow, color=black] ($(target_detection.east) - (0, 2*\varrowshift)$) -- ($(lighting_trajectory_generator.west) - (0, 2*\varrowshift)$) node[midway, below, shift={(0.0, -0.0*\vlabelshift)}] {\small $T_T$} ;
    \draw [arrow, color=black] ($(target_detection.east) - (0, -2*\varrowshift)$) -| ($(target_detection.east) + (0.15, 1.0)$) |- ($(cinematographic_trajectory_generator.west) - (0, -0*\varrowshift)$) node[midway, left, shift={(0.0, -10.8*\vlabelshift)}] {\small $T_T$} ;
    \draw [arrow, color=black] ($(cinematographic_trajectory_generator.east) - (0, -0*\varrowshift)$) -| ($(path_generator.north) - (\harrowshift, -0*\varrowshift)$) node[midway, above, shift={(-2.5*\hlabelshift, 0.0*\vlabelshift)}] {\small $D_L$} ;
    \draw [arrow, color=black] ($(lighting_trajectory_generator.east) - (0, -2*\varrowshift)$) -| ($(path_generator.south) - (\harrowshift, -0*\varrowshift)$) node[midway, above, shift={(-5.0*\hlabelshift, 0.0*\vlabelshift)}] {\small $D_F$} ;
    \draw [arrow, color=black] ($(trajectory_optimizer.south) - (0, -0*\varrowshift)$) |- ($(lighting_trajectory_generator.east) - (0, 2*\varrowshift)$) node[midway, below, shift={(-17.0*\hlabelshift, 0.0*\vlabelshift)}] {\small $T_L$, $T_F$ } ;
    \draw [arrow, color=black] ($(path_generator.east) - (0, -2*\varrowshift)$) -- ($(safe_corridor_generator.west) - (0, -2*\varrowshift)$) node[midway, above, shift={(0.0, -0.8*\vlabelshift)}] {\small $P_L$ } ;
    \draw [arrow, color=black] ($(path_generator.east) - (0, 2*\varrowshift)$) -- ($(safe_corridor_generator.west) - (0, 2*\varrowshift)$) node[midway, below, shift={(0.0, 0.4*\vlabelshift)}] {\small $P_F$ } ;
    \draw [arrow, color=black] ($(safe_corridor_generator.east) - (0, -2*\varrowshift)$) -- ($(trajectory_optimizer.west) - (0, -2*\varrowshift)$) node[midway, above, shift={(0.0, -0.8*\vlabelshift)}] {\small $S_L$ } ;
    \draw [arrow, color=black] ($(safe_corridor_generator.east) - (0, 2*\varrowshift)$) -- ($(trajectory_optimizer.west) - (0, 2*\varrowshift)$) node[midway, below, shift={(0.0, 0.4*\vlabelshift)}] {\small $S_F$ } ;
    \draw [arrow, color=black] ($(trajectory_optimizer.east) - (0, -2*\varrowshift)$) -- ($(trajectory_tracker.west) - (0, -2*\varrowshift)$) node[midway, above, shift={(0.0, -0.8*\vlabelshift)}] {\small $T_L$ } ;
    \draw [arrow, color=black] ($(trajectory_optimizer.east) - (0, 2*\varrowshift)$) -- ($(trajectory_tracker.west) - (0, 2*\varrowshift)$) node[midway, below, shift={(0.0, 0.4*\vlabelshift)}] {\small $T_F$} ;
    \draw [arrow, color=black] ($(trajectory_optimizer.south) - (\harrowshift, 0*\varrowshift)$) |- ($(trajectory_optimizer.south) + (-0.5, -\varrowbend)$) -| ($(path_generator.south) - (-\harrowshift, -0*\varrowshift)$) node[midway, below, shift={(11.4*\hlabelshift, 0.7*\vlabelshift)}] {\small $T_L$, $T_F$ } ;
    \draw [arrow, color=black] ($(trajectory_optimizer.south) - (\harrowshift, 0*\varrowshift)$) |- ($(trajectory_optimizer.south) + (-0.5, -\varrowbend)$) -| ($(safe_corridor_generator.south) - (-0*\harrowshift, -0*\varrowshift)$) node[midway, above, shift={(0.0, 4.8*\varrowshift)}] {\small } ;
   \draw [arrow, color=black] ($(map.south) - (0.6, -0.1)$) |- ++(-0.0, -0.4*\varrowbend) -| ($(path_generator.north) - (-\harrowshift, 0)$) node[midway, right, shift={(0.0, 0)}] {\small };
   \draw [arrow, color=black] ($(map.south) - (0.6, -0.1)$) |- ++(-0.0, -0.4*\varrowbend) -| ($(safe_corridor_generator.north) - (\harrowshift, 0)$) node[midway, right, shift={(0.0, 0)}] {\small };
   \draw [arrow, color=black] ($(dynamic_constraints.north) - (0, 0.1)$) |- ++(-0.0, 0.5*\varrowbend) -| ($(path_generator.north) - (0.0, 0)$) node[midway, right, shift={(0.0, 0)}] {\small };
    \draw [arrow, color=black] ($(dynamic_constraints.south) - (0, -0.1)$) -- ($(trajectory_optimizer.north) - (0, -0*\varrowshift)$) node[midway, above, shift={(0.0, 4.8*\varrowshift)}] {\small } ;

  \end{pgfonlayer}
    %%{ background
    
    % \begin{pgfonlayer}{background}
    %   \path (human_director.west |- human_director.north)+(-0.20,0.5) node (a) {};
    %   \path (cinematographic_trajectory_generator.south -| cinematographic_trajectory_generator.east)+(+0.3,-0.3) node (b) {};
    %   \path[fill=blue!10,rounded corners, draw=black!70, densely dotted]
    %   (a) rectangle (b);
    % \end{pgfonlayer}
    % \node [rectangle, above of=human_director, node distance=1.7em, shift={(0.7,0.05)}] (text_control) {\footnotesize High-level planning};
    
    \begin{pgfonlayer}{background}
      \path (path_generator.west |- path_generator.north)+(-0.12,0.12) node (a) {};
      \path (trajectory_tracker.south -| trajectory_tracker.east)+(+0.12,-0.12) node (b) {};
      \path[fill=blue!05,rounded corners, draw=black!70, densely dotted]
      (a) rectangle (b);
    \end{pgfonlayer}
    %%}

  \end{tikzpicture}
 \caption{The architecture of the proposed system. $C_s$ and $C_l$ represent the desired type of cinematographic shot and lighting configuration specified by a human director;
   $T_T$ is the target estimated trajectory;
   $D_L$, $D_F$ are reference trajectories for the leader UAV and the follower UAVs, respectively;
   $P_L$, $P_F$ are collision-free paths generated along the desired trajectories;
   $S_L$, $S_F$ are safe corridors along the collision-free paths;
   and $T_L$, $T_F$ are optimized trajectories for the camera and lighting UAVs, respectively.
   The modules enclosed in the blue rectangle run on both types of UAVs.
   }
   \vspace{-0.4cm}
  \label{fig:system_architecture}
\end{figure*}
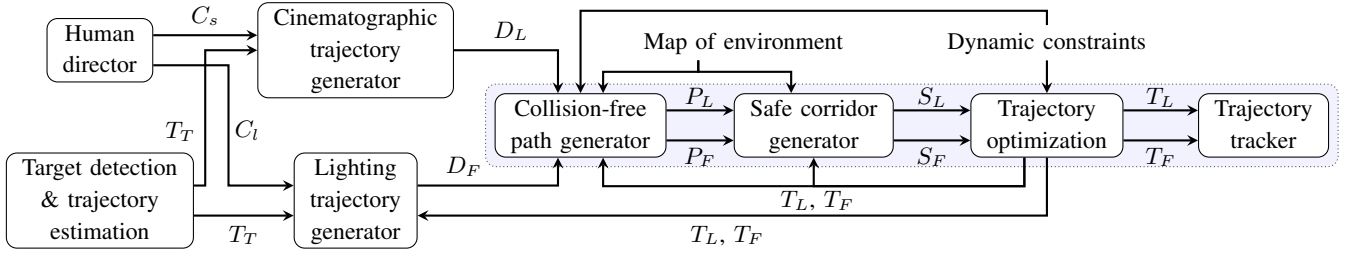

\autoref{fig:system_architecture} depicts the architecture of the entire system. The leader UAV carries a camera for filming while several others carry light sources to provide proper illumination. A human director specifies the cinematographic parameters for the scene. These parameters include the shot type (i.e., the camera motion relative to the target), the camera shooting angle for the leader, and the desired lighting angles for the followers. This information, together with an estimation of the target trajectory, is used to generate reference trajectories for the UAVs (\autoref{sec:initial_path}). These initial trajectories do not consider obstacle avoidance, but only cinematographic aspects. The leader attempts to execute the commanded shot smoothly, whereas the followers maintain a surrounding formation with the desired lighting angles.

Safety is ensured by integrating information from a local map for collision avoidance (\autoref{sec:safe_corridor_generation}). Firstly, a collision-free path is generated for each UAV using the map and the initial cinematographic trajectories as guidelines. Then, a safe corridor along each of these paths is computed, consisting of a set of obstacle-free polyhedrons generated by the convex decomposition of free space (see~\autoref{fig:path_conversion}).
Finally, the UAV trajectories are obtained as a result of a trajectory optimization process that computes dynamically feasible trajectories inside each safe corridor (\autoref{sec:sec:final_trajectory_generation}). 
Inter-UAV collision avoidance is achieved by including the team-mates planned trajectories as obstacles in the map. 

The entire pipeline shown in \autoref{fig:system_architecture} (except for the \emph{Human director} component) runs on board each UAV in a receding horizon manner. This enables the online planning to react properly to changes in the behavior of the target being filmed, as well as to malfunctioning team-members or previously unseen obstacles.
Note that either the \emph{Cinematographic trajectory generator} or the \emph{Lighting trajectory generator} is activated on each UAV, depending on whether it carries a camera or light. 
The component for trajectory tracking on each UAV is the low-level control pipeline described in~\cite{baca2020mrs}.

\section{AUTONOMOUS AERIAL CINEMATOGRAPHY}
\label{sec:method}

In this section, we begin by detailing the UAV dynamic model (\autoref{sec:dynamic_model}). Then, we describe our procedure to generate optimal and safe trajectories for each UAV (Sections~\ref{sec:initial_path}, \ref{sec:safe_corridor_generation}, and \ref{sec:sec:final_trajectory_generation}). 
Lastly, we explain how the orientation of a UAV is controlled (\autoref{sec:orientation}).

\subsection{Multi-rotor aerial vehicle dynamic model}\label{sec:dynamic_model}

An independent trajectory tracker~\cite{baca2020mrs} for UAV attitude control is used, which allows for planning with a simplified positional dynamic UAV model. In addition, the orientation of the camera or light source onboard (depending on the UAV) needs to be modelled. We assume the existence of a gimbal mechanism to compensate angle deviations due to changes in UAV attitude. Therefore, it is assumed that camera roll is negligible and we only control pitch and heading. Since the heading of a multi-rotor vehicle can be controlled independently of its position, we fix the relative position between the camera/light and the UAV to always point forward and control its heading through the UAV heading.    
The positional part of the dynamic model is defined as a linear double integrator:
\begin{equation}\label{eq:dynamics_p}
  \begin{split}
    \dot{\mathbf{p}} &= \mathbf{v},\\
    \dot{\mathbf{v}} &= \mathbf{a},
  \end{split}
\end{equation}
\noindent where $\mathbf{p} = [p_x\ p_y\ p_z]^T\in\mathbb{R}^3$ is the UAV position, $\mathbf{v} = [v_x\ v_y\ v_z]^T\in\mathbb{R}^3$ the linear velocity, and $\mathbf{a} = [a_x\ a_y\ a_z]^T \in\mathbb{R}^3$ the linear acceleration.
The orientation of the camera/light may be modelled similarly: 
\begin{equation}\label{eq:dynamics_o}
  \begin{split}
    \dot{\mathbf{o}} &= \boldsymbol{\omega}, \\
    \dot{\boldsymbol{\omega}} &= \boldsymbol{\theta},
  \end{split}
\end{equation}
\noindent where $\mathbf{o} = [\varphi\ \xi]^T$ represents an orientation with respect to a global frame given by its heading and pitch angles, $\boldsymbol{\omega} \in\mathbb{R}^2$ are the corresponding angular rates, and $\boldsymbol{\theta} \in\mathbb{R}^2$ the angular accelerations.
For the description of the proposed method, we define a full positional state of the UAV $\mathbf{x}_p = [\mathbf{p}^T\ \mathbf{v}^T]^T \in \mathbb{R}^6$, a vector of positional control inputs $\mathbf{u}_p = \mathbf{a}$, an orientation state $\mathbf{x}_o = [\mathbf{o}^T\ \boldsymbol{\omega}^T]^T \in \mathbb{R}^4$, and a vector of orientation control inputs $\mathbf{u}_o = \boldsymbol{\theta}$. 

\subsection{Generation of reference trajectories}\label{sec:initial_path}

The first step of our method for trajectory planning is to generate a reference trajectory $D_j$ for each UAV $j$. The problem complexity is alleviated by removing collision avoidance constraints and focusing only on cinematographic aspects. For the filming UAV, the objective is to reach a position relative to the target as provided by the shot type $C_s$, while minimizing changes in the camera angle to produce pleasant images. A specific camera shooting angle $\psi_d$ over the target needs to be maintained. 
The following non-linear optimization problem is formulated\footnote{For simplicity of description, $\mathbf{x}:=\mathbf{x}_p$, and $\mathbf{u}:=\mathbf{u}_p$. We use the Runge-Kutta method for numerical integration.} for the filming UAV:
\begin{align} 
\label{eq:formulation_initial}
\underset{\begin{subarray}{c}
  \mathbf{u_0},\dots,\mathbf{u_{N-1}}
  \end{subarray}}{\text{minimize}} \ &  \sum_{k=1}^{N} ( ||\mathbf{u}_{k-1}||^2 + \alpha_1 J_{\psi,k}) + \alpha_2 J_N,& \\ 
  \text{subject to} \ \mathbf{x}_0 &= \mathbf{x}',& \label{eq:formulation_initial_initial} \tag{\ref{eq:formulation_initial}.a} \\ 
\mathbf{x}_{k+1} &= \text{f}_p(\mathbf{x}_{k},\mathbf{u}_{k}) \quad\;\: \forall k \in \{0,\dots, N-1\}, \label{eq:formulation_initial_dyn}
\tag{\ref{eq:formulation_initial}.b} \\
\mathbf{v}_{min} &\leq \mathbf{v}_{k} \leq \mathbf{v}_{max} \;\;\;\: \forall k \in \{1,\dots, N\},& \label{eq:formulation_initial_vel} \tag{\ref{eq:formulation_initial}.c} \\
\mathbf{u}_{min} &\leq \mathbf{u}_{k} \leq \mathbf{u}_{max}\;\;\;\: \forall k \in \{0,\dots, N-1\}, \label{eq:formulation_initial_input}\tag{\ref{eq:formulation_initial}.d} \\
q_{z,min} &\leq q_{z,k} \qquad \qquad \,  \forall k \in \{1,\dots, N\},&
\label{eq:formulation_initial_qz} \tag{\ref{eq:formulation_initial}.e}
\end{align}
where $\text{f}_p(\cdot)$ represents the positional part of the dynamic model defined in \autoref{sec:dynamic_model}; $\mathbf{v}_{min}$, $\mathbf{v}_{max}$ are velocity limitations; and $\mathbf{u}_{min}$, $\mathbf{u}_{max}$  control inputs limitations. 

\begin{figure}[htb]
    \centering
    \includegraphics[width=\columnwidth]{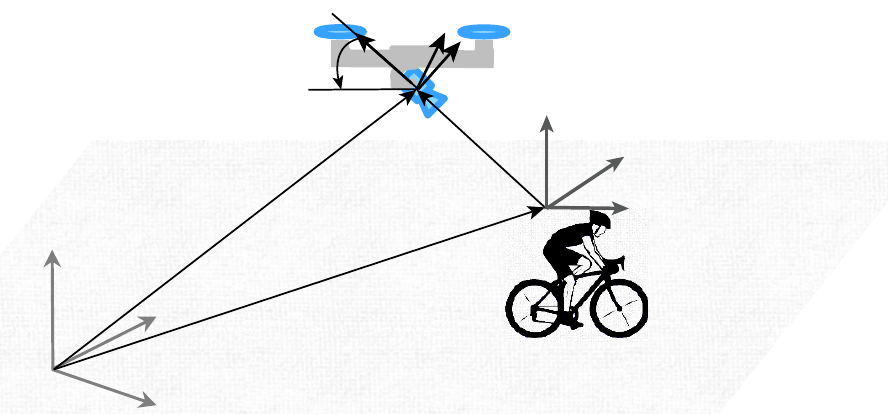}
    \put(-170,97){$\psi_d$}
    \put(-105,100){ Leader}
    \put(-204,5){ Global frame}
    \put(-100,6){Target}
    \put(-178,65){$p_{L}$}
    \put(-110,75){$q$}
    \put(-145,49){$p_T$}
    \caption{Reference frames and camera shooting angle. The origins of the camera and UAV frames coincide. 
    }
    \label{fig:cinematography_term_scheme}
\end{figure}
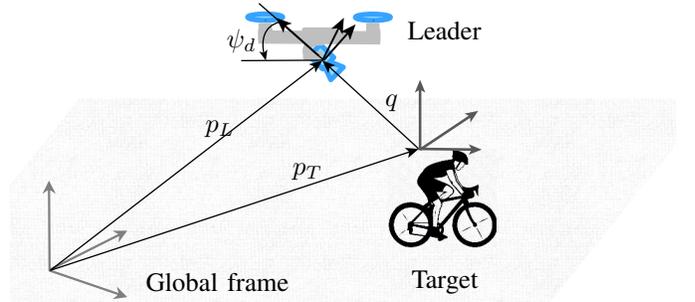

The first two terms in the cost function pursue smooth trajectories by penalizing UAV accelerations and reducing gimbal movements.
The director specifies an aesthetic objective through the desired camera shooting angle $\psi_d$ to film the target (see \autoref{fig:cinematography_term_scheme}). Emphasis is given on positioning the UAV to keep this angle constant without moving the gimbal. In doing so, the angular changes in the gimbal are reduced to favor less jerky camera motion and therefore, pleasant footage. 
In order to define $J_\psi$, the relative position between the UAV camera and the target is introduced as: 
\begin{equation}
 \mathbf{q} = \begin{bmatrix}
        q_x & q_y & q_z
    \end{bmatrix}^T
= \mathbf{p}_L - \mathbf{p}_T.   
\end{equation}
Then, we define $J_\psi$ as: 
\begin{equation}
J_{\psi,k}= \left(\text{tan}(\psi_d) - \frac{q_{z,k}}{\sqrt{q_{x,k}^2+q_{y,k}^2}}\right)^2.
\end{equation}
The target position is predicted within the time horizon with a motion model (a constant speed model in our experiments). Prediction errors are tackled by recomputing UAV trajectories with a receding horizon.
By minimizing the previous cost, we implicitly minimize variations in camera pitch angle as the relative pitch with respect to the target is kept constant. Moreover, the camera heading corresponds with the UAV heading, whose variations are also smoothed as explained in \autoref{sec:sec:final_trajectory_generation}. Therefore, the idea is to generate UAV trajectories where the gimbal only needs to move slightly to compensate for small disturbances. 

The terminal cost $J_N=||\mathbf{x}_{xy,d}-\mathbf{x}_{xy,N} ||^2$ guides the UAV to a desired state imposed by the shot type, e.g., at a certain distance beside the target's final position in a lateral shot. 
Note that a final UAV height is not imposed, as we want the planner to compute the optimal $p_z$ to maintain the camera shooting angle commanded by the director. Lastly, the constraint~\eqref{eq:formulation_initial_qz} establishes a minimum distance above the target for safety purposes. 

\begin{figure}[htb]
    \centering
    \input{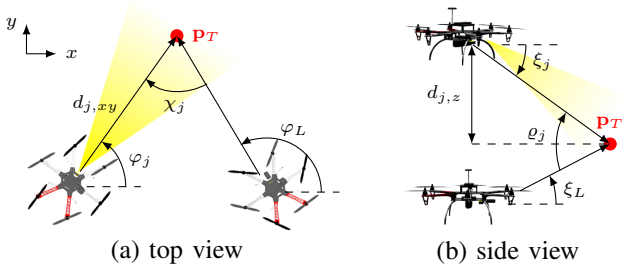}
    \vspace{-0.6cm}
    \caption{The leader-follower scheme defined by~\eqref{eq:leader_follower_equations}.}
    \label{fig:formation_scheme}
\end{figure}

The reference trajectories for the lighting UAVs are computed to achieve a desired leader-follower formation around the target. The desired position of the followers is influenced by the corresponding leader position $\mathbf{p}_L$ and camera orientation $\mathbf{o}_L$, the target position $\mathbf{p}_T$, the desired lighting angles of $j$-th light $\chi_j$ and $\varrho_j$, and the desired distance of the light to the target $d_j$.
The desired position of $j$-th follower $\mathbf{p}_j$ is then given by the equation:
\begin{equation}\label{eq:leader_follower_equations}
\begin{split}
  \mathbf{p}_j &= \mathbf{p}_T + d_j \begin{bmatrix}
    -\cos(\varphi_j)\cos(\xi_j) \\
    -\sin(\varphi_j)\cos(\xi_j) \\
    \sin(\xi_j)
    \end{bmatrix},
\end{split}
\end{equation}
where $\varphi_j = \varphi_L+\chi_j$ and $\xi_j = \xi_L+\varrho_j$ are desired lighting angles relative to the camera's optical axis (see~\autoref{fig:formation_scheme}). 
To avoid jumps in the desired followers' positions caused by quick changes in the target position (e.g., due to a transition to a new target), a virtual target, located in front of the camera at a certain distance along its optical axis, is used.
The position of this virtual target is given by: 
\begin{equation}\label{eq:virtual_ooi}
  \mathbf{p}_v = \mathbf{p}_L + d_v \begin{bmatrix}
    \cos(\varphi_L)\cos(\xi_L) \\
    \sin(\varphi_L)\cos(\xi_L) \\
    \sin(\xi_L)
    \end{bmatrix},
\end{equation}
where $d_v$ is the desired distance between the virtual target and the camera center and $\mathbf{p}_v$ denotes the virtual target position. Substituting position $\mathbf{p}_v$ for $\mathbf{p}_T$ in~\eqref{eq:leader_follower_equations}, a more consistent formation scheme is acquired, where less aggressive maneuvers are required; and the lighting always focuses on the scene in front of the camera, which is relevant in obtaining pleasant videos.

\subsection{Generation of safe corridors}\label{sec:safe_corridor_generation}

The initial reference trajectories are computed without considering obstacles. They are, therefore, used as seed to generate a safe corridor $S_j$ for each UAV $j$ where collision-free trajectories can then be computed.
Firstly, we convert each trajectory $D_j$ into a collision-free path $P_j$. We iterate over each of the $N$ waypoints in $D_j$ and add it directly to $P_j$ if it is collision-free. Otherwise, we label the previous collision-free waypoint as $A$ and keep moving along $D_j$ until we find the next collision-free waypoint $B$. Then, we try to find an alternative collision-free path from $A$ to $B$, to be appended to $P_j$ and continue iterating. For that alternative path, we use the \emph{Jump Point Search} (JPS) algorithm introduced in~\cite{jpsHarabor2011, harabor2014} and extended to 3D in~\cite{Liu2017}. A real-time performance is ensured by introducing a timeout for the JPS path search.

\begin{figure}[htb]
    \vspace{0.1cm}
    \centering
    \includegraphics[width=\columnwidth]{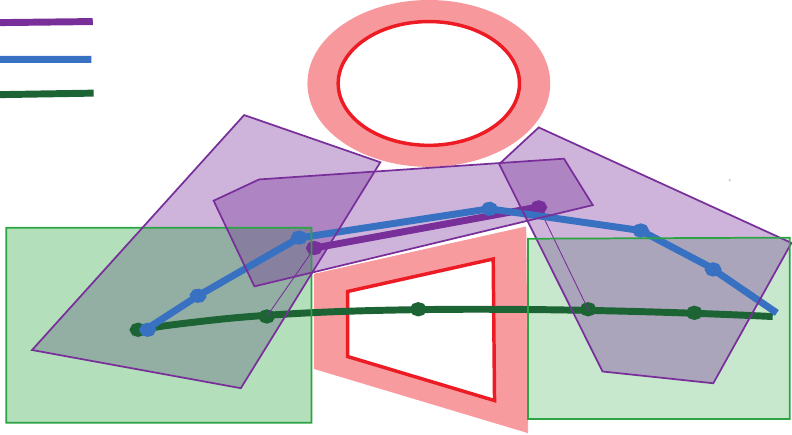}
    \put(-134,108){\color{red} Obstacle}
    \put(-136,29){\color{red} Obstacle}
    \put(-218,106){\small Initial trajectory}
    \put(-218,118){\small Final trajectory}
    \put(-218,130){\small Collision free  path}
    \caption{The safe corridor generation process. The initial reference trajectory (green) is converted into a collision-free path (purple), and the obstacle-free polyhedrons are generated along this path. The final optimized trajectory within the safe corridor is also shown (blue). We inflate the obstacles for safety purposes (light red).}
    \label{fig:path_conversion}
    \vspace{-0.55cm}
\end{figure}
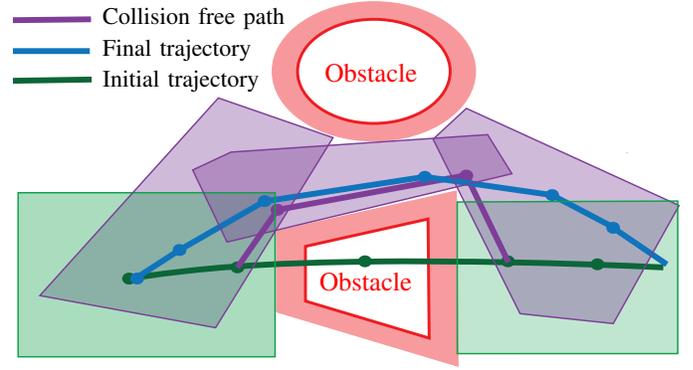

If the JPS algorithm fails to find a path within the given timeout from $A$ to $B$, we run it again to connect $A$ directly to the last waypoint in $D_j$ (let this waypoint be $C$). If this is not found either, we append to $P_j$ the path to the node closest to $C$ from all those expanded during the JPS search.   
Once completed, $P_j$ consists of an arbitrary number of points equal to or greater than $N$. Since $P_j$ is used for the generation of the safety corridors for particular points in $D_j$, it is post-processed so that $|P_j| = |D_j| = N$. $P_j$ is sampled so that the waypoint distribution is close to the initial points in $D_j$. Since these collision-free paths are used as a guide for trajectory optimization in subsequent steps, the distance sampling step $d_{s}$ is limited to help avoid the dynamic infeasibility of the final trajectories.
If the sampled $P_j$ consists of more than $N$ waypoints, the overflowing points are discarded for the subsequent steps of the trajectory optimization process.  
The process to create a collision-free path $P_j$ and its corresponding safe corridor $S_j$ is illustrated in~\autoref{fig:path_conversion}.

Safe corridors are generated around the collision-free paths with a prefixed initial position of the UAV (i.e., $N+1$ waypoints), using a map of the environment represented by a point cloud $O_{pcl}$ and the convex decomposition method proposed in~\cite{Liu2017}.
This method is based on an iterative procedure for the generation of polyhedrons. 
It begins by inflating an ellipsoid aligned with each path segment.
In the next step, tangent planes are constructed at the contact points between the ellipsoid and any obstacles.  
Afterwards, all points lying behind this plane are removed from $O_{pcl}$. Yet again, the next iteration starts by inflating the ellipsoid up to the nearest point in $O_{pcl}$. This procedure is terminated if there are no remaining points in $O_{pcl}$. The generated tangent planes define an obstacle-free polyhedron $\mathcal{P}$ enclosing the corresponding path segment and the set of all polyhedrons along the path constitutes the safe corridor.     

\subsection{Trajectory optimization}
\label{sec:sec:final_trajectory_generation}
Given a collision-free path $P$ and its corresponding safe corridor $S$, a final optimal trajectory is computed through a QP problem in receding horizon. 
The particular optimization task~\footnote{For simplicity of description, $\mathbf{x}:=\mathbf{x}_p$, and $\mathbf{u}:=\mathbf{u}_p$.} attempts to track a desired trajectory $\mathbf{p}_d$ corresponding to the reference trajectory $D_j$:
\begin{align}
\label{eq:formulation_follower}
\underset{\begin{subarray}{c}
  \mathbf{u_0},\dots,\mathbf{u_{N-1}}
  \end{subarray}}{\text{minimize}} \ &  \sum_{k=1}^{N} (||\mathbf{p}_{d,k} - \mathbf{p}_k||^2 + \beta ||\mathbf{u}_{k-1}||^2), \\ 
  \text{subject to} \ \mathbf{x}_0 &= \mathbf{x}', \label{eq:formulation_follower_init} \tag{\ref{eq:formulation_follower}.a} \\ 
  \mathbf{x}_{k+1} &= \text{f}_p(\mathbf{x}_{k},\mathbf{u}_{k}) \quad  \forall k \in \{0,\dots, N-1\}, \label{eq:formulation_follower_dyn} \tag{\ref{eq:formulation_follower}.b} \\
  \mathbf{v}_{min} &\leq \mathbf{v}_{k} \leq \mathbf{v}_{max} \; \; \forall k \in \{1,\dots, N\}, \label{eq:formulation_follower_vel} \tag{\ref{eq:formulation_follower}.c}  \\
  \mathbf{u}_{min} &\leq \mathbf{u}_{k} \leq \mathbf{u}_{max} \; \:\forall k \in \{0,\dots, N-1\}, \label{eq:formulation_follower_input} \tag{\ref{eq:formulation_follower}.d} \\
  \mathbf{p}_{k} &\in \mathcal{P}_k \qquad \qquad\, \forall k \in \{1,\dots, N\}, \label{eq:formulation_follower_polyhedron} \tag{\ref{eq:formulation_follower}.e}
\end{align}
where $\text{f}_p(\cdot)$ represents the positional part of a dynamic model defined in \autoref{sec:dynamic_model}; $\mathbf{v}_{min}$, $\mathbf{v}_{max}$ are velocity limitations; $\mathbf{u}_{min}$, $\mathbf{u}_{max}$ control inputs limitations; and $\mathcal{P}_k$ is a convex polyhedron representing a free space associated with $k$-th transition point. 
The last constraint ensures a safe resulting trajectory without collisions. 
Given that the constraint~\eqref{eq:formulation_follower_polyhedron} can be decoupled in a set of linear constraints, the problem becomes a quadratic convex program. 

The optimization formulation is the same for both the leader and follower UAVs. However, there are a couple of relevant differences. First, the desired reference trajectories are computed in a different manner, following either filming or lighting criteria (see \autoref{sec:initial_path}). Second, the followers encode mutual-collision avoidance through constraint~\eqref{eq:formulation_follower_polyhedron}.  
To prevent negative effects on the cinematographic quality of the performed shot, the entirety of mutual collision avoidance is left to the followers.
A fixed priority scheme is defined for the UAVs, and the occupied space $O_{pcl}$ of each follower is updated with the current planned trajectories from the leader and other followers of a higher priority. $O_{pcl}$ is updated with spherical objects of the desired collision avoidance radius at each waypoint of the UAV trajectories to be avoided. A similar procedure is followed to incorporate the target's predicted trajectory (also for the leader in this case). To hold with real-time performance, the occupied space $O_{pcl}$ is assumed static for a given horizon time, but it is updated at each planning step, accommodating all static and dynamic obstacles.

Another crucial issue for the applications of multi-UAV cinematography is how to prevent other UAVs from appearing in the \emph{Field of View} (FoV) of the filming UAV. However, including this in the optimization task as either a constraint or a cost term can remarkably increase the complexity of the problem. We considered including the FoV of the leader camera as an obstacle in the local maps of the followers, so that they may avoid it. Even so, relatively small changes in camera orientation could result in significant changes in the map representation and lead to unstable planned trajectories. 
Therefore, the camera's FoV is avoided by the lighting UAVs only through penalizing deviations from the desired trajectories $\mathbf{p}_d$. Thus, FoV avoidance is mostly determined by the choice of lighting parameters that describe the desired formation. 

Finally, occlusions caused by obstacles in the FoV of the camera or the lights are also a relevant aspect when filming. 
Occlusion throughout a significant part of the shot renders the shot useless, and in the case of onboard detection of the target, it also disables target following.
However in most cases, occlusions are temporary and avoiding them is always a trade-off between significant deviation from the desired cinematographic trajectory and having part of the video occluded.
In this work, the trajectories are generated so that they are close to the desired cinematographic shots specified by a director. The possible occlusions have to be resolved by redefining the shot to be performed.

\subsection{Orientation control} 
\label{sec:orientation}

In this application, both the camera and the light sources need to always be pointing at the filmed target. 
Hence, their desired orientation is given by:
\begin{equation}\label{eq:desired_orientation}
  \mathbf{o}_d =\begin{bmatrix}
    \varphi_d \ \xi_d
    \end{bmatrix}^T =  
  \begin{bmatrix}
    \text{arctan}(q_y, q_x)\ \sin\left(\frac{q_z}{||q||}\right)
    \end{bmatrix}^T.
\end{equation}

Orientation control is also formulated as a constrained quadratic optimization problem in receding horizon in order to achieve smoother orientation changes. 
For simplicity of description, $\mathbf{x}:=\mathbf{x}_o$ and $\mathbf{u}:=\mathbf{u}_o$ in the following problem formulation:
\begin{alignat}{3} 
\label{eq:formulation_orientation}
\underset{\begin{subarray}{c}
  \mathbf{u_{0}},\dots,\mathbf{u_{N-1}}
  \end{subarray}}{\text{minimize}} \ &  \sum_{k=1}^{N} (||\mathbf{o}_{d,k} - \mathbf{o}_k||^2 + \gamma ||\mathbf{u}_{k-1}||^2), & \\ 
  \text{subject to} \ \mathbf{x}_{0} &= \mathbf{x}', & \label{eq:formulation_orientation_init} \tag{\ref{eq:formulation_orientation}.a} \\ 
  \mathbf{x}_{k+1} &= \text{f}_o(\mathbf{x}_{k},\mathbf{u}_{k}) \quad\: \forall k \in \{0,\dots, N-1\}, & \label{eq:formulation_orientation_dyn} \tag{\ref{eq:formulation_orientation}.b}\\
  \boldsymbol{\omega}_{min} &\leq \boldsymbol{\omega}_{k} \leq \boldsymbol{\omega}_{max} \;\: \forall k \in \{1,\dots, N\}, & \label{eq:formulation_orientation_vel} \tag{\ref{eq:formulation_orientation}.c} \\
  \xi_{min} &\leq \xi_{k} \leq \xi_{max} \;\;\;\; \forall k \in \{1,\dots, N\}, & \label{eq:formulation_orientation_xi} \tag{\ref{eq:formulation_orientation}.d}\\
  \mathbf{u}_{min} &\leq \mathbf{u}_{k} \leq \mathbf{u}_{max} \;\;\, \forall k \in \{0,\dots, N-1\}, & \label{eq:formulation_orientation_input} \tag{\ref{eq:formulation_orientation}.e}
\end{alignat}
where $\text{f}_o(\cdot)$ represents the orientation aspect of the dynamic model defined in~\autoref{sec:dynamic_model}; $\mathbf{\omega}_{min}$, $\mathbf{\omega}_{max}$ are limitations on the angular velocities; $\mathbf{u}_{min}$, $\mathbf{u}_{max}$ control inputs limitations; and $\xi_{min}$, $\xi_{max}$ represent hardware limitations of the gimbal to adjusting pitch angles. 
The heading and pitch angles of the camera or light can be controlled independently. Thus, Problem~\eqref{eq:formulation_orientation} was decoupled into two simpler problems. The optimal solution for each problem can be found analytically with a standard framework for linear MPC (\emph{Model Predictive Control}). 

\section{EXPERIMENTAL EVALUATION}
\label{sec:experimental_evaluation}

In this section, experimental results are presented to demonstrate the performance of our method for multi-UAV trajectory planning. 
We have assessed that the proposed method is capable of computing smooth cinematographic trajectories in real-time. Additionally, we have evaluated that the trajectories of the follower UAVs which provide lighting for the target are capable of complying with formation constraints to improve the quality of the shot. The safety of our method has also been proved through experiments in the presence of multiple obstacles. 

\subsection{Experimental setup}

We implemented our architecture described in~\autoref{sec:system_overview} in C++ using the ROS framework. The ACADO Toolkit~\cite{Houska2011a} was used to solve the optimization problems. We conducted software-in-the-loop simulations using Gazebo to simulate physics and to equip the UAVs with a camera and lights. To solve the optimization problems, a horizon length of \SI{8}{\second} and a time step of \SI{0.2}{\second} were chosen. The cinematographic parameters were set to $\psi_d=6^\circ$ and $q_{z,min} = \SI{0.5}{\meter}$. The maximum distance sampling step was set to $d_{s,max}$ = \SI{0.5}{m}.

\subsection{Simulation - Cinematography trajectories}

The objective of this simulation was twofold: to demonstrate how the method computes smoother camera trajectories for the leader UAV while complying with cinematographic aspects, and how the trajectories of the followers keep with formation constraints to light the target properly. We simulated a human worker performing a maintenance operation on a transmission tower while monitored by a team of three UAVs (one filming and two lighting the worker). While the worker approached and climbed the tower, the system was commanded to perform a lateral shot followed by a sequence of fly-over shots.

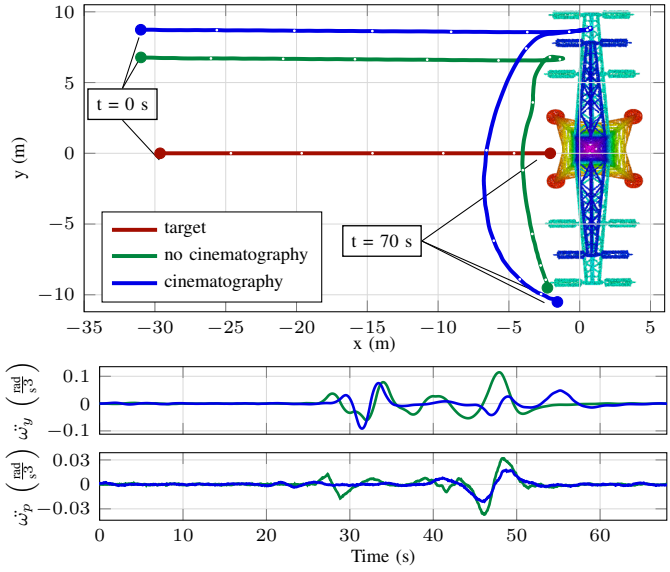
\begin{figure}
    \centering
    \vspace{-0.2cm}
    \hspace*{-0.6cm}
  \subfloat {
    % \pgfplotsset{compat=1.9}

\definecolor{color_red}{HTML}{A30D00}
\definecolor{color_green}{rgb}{0, .522, .243}
\definecolor{color_blue}{rgb}{0, 0, .9}

\begin{tikzpicture}[font=\scriptsize, trim axis right]

  %%{ start and end box
  \coordinate (start_from) at (0.51,2.55);
  \coordinate (start_to) at (0.73,3.35);
  \coordinate (start_to3) at (0.73,3.65);
  \coordinate (start_to2) at (1.00,2.02);
  \coordinate (end_from) at (3.96,0.73);
  \coordinate (end_to) at (5.98,0.28);
  \coordinate (end_to2) at (6.02,2.02);
  \coordinate (end_to3) at (6.12,0.12);
  
  % R1
  \node[legend_box, above, line width=0.5pt] (start) at (start_from) {
    t = 0 s
  };
  \draw (start.north) -- (start_to);
  \draw (start.south) -- (start_to2);
  \draw (start.north) -- (start_to3);
  
  \node[legend_box, above, line width=0.5pt] (end) at (end_from) {
    t = 70 s
  };
  \draw (end.east) -- (end_to);
  \draw (end.east) -- (end_to2);
  \draw (end.east) -- (end_to3);
  
  %%}

  \begin{axis}[
    name=top,
    unit vector ratio*=1 1 1,
    width=1.05\columnwidth,
    % height=0.48\columnwidth,
    grid=both,
    grid style={draw=gray!30,line width=.1pt},
    xlabel= x (m),
    ylabel= y (m),
    set layers,
    xmin=-35, xmax=6,
    ymin=-11.2,ymax=10.5,
    legend style={font=\scriptsize},
    legend pos=south west,
    legend cell align={left},
    x label style={at={(axis description cs:0.5,-0.05)},anchor=north},
    y label style={at={(axis description cs:-0.08,.5)},rotate=0,anchor=south},
    % axis equal image,
    % ytick={0,1,2,3,4},
    ]

    %%{ add trajectories 
    
    \addplot[smooth, color=color_red, line width=1.5pt] table[y=x, x=y] {fig/tikz_data/top_view_target2.txt};  
    \addplot[smooth, color=color_green, line width=1.5pt] table[y=x, x=y] {fig/tikz_data/top_view_no_cin2.txt};  
    \addplot[smooth, color=color_blue, line width=1.5pt] table[y=x, x=y] {fig/tikz_data/top_view_cin2.txt};  
    
    %%}

    %%{ add white trajectory markers
    
    \addplot[thick, color=blue, on layer=axis background]
    graphics[xmin=-30,ymin=-12,xmax=9,ymax=12] {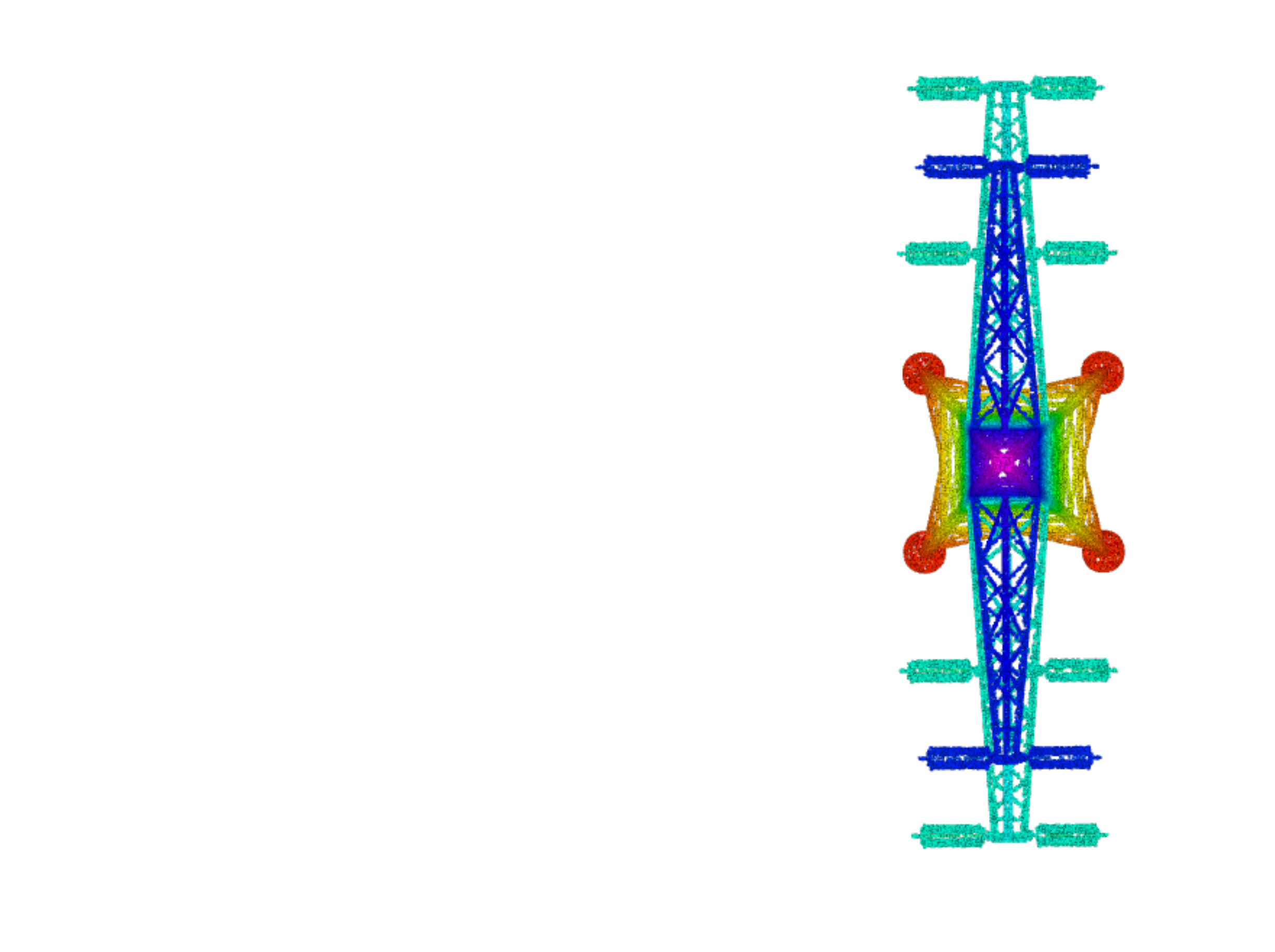};
    \addplot [color=white, only marks, mark=*, mark size=0.3pt, solid, each nth point={100}, forget plot] table[y=x, x=y] {fig/tikz_data/top_view_target2.txt};
    \addplot [color=white, only marks, mark=*, mark size=0.3pt, solid, each nth point={100}, forget plot] table[y=x, x=y] {fig/tikz_data/top_view_cin2.txt};
    \addplot [color=white, only marks, mark=*, mark size=0.3pt, solid, each nth point={100}, forget plot] table[y=x, x=y] {fig/tikz_data/top_view_no_cin2.txt};
    
    %%}

  \addlegendentry{target} 
  \addlegendentry{no cinematography} 
  \addlegendentry{cinematography} 

  %%{ Start and end markers
  
    \addplot [color=color_green, only marks, mark=*, mark size=2.0pt, solid, each nth point={100000}, forget plot] table[y=x, x=y] {fig/tikz_data/top_view_no_cin2.txt};
    \addplot [color=color_blue, only marks, mark=*, mark size=2.0pt, solid, each nth point={100000}, forget plot] table[y=x, x=y] {fig/tikz_data/top_view_cin2.txt};
    \addplot [color=color_red, only marks, mark=*, mark size=2.0pt, solid, each nth point={100000}, forget plot] table[y=x, x=y] {fig/tikz_data/top_view_target2.txt};
  
    \addplot [color=color_red, only marks, mark=*, mark size=2.0pt, solid, forget plot] coordinates {(-2.08,0.0)};
    \addplot [color=color_blue, only marks, mark=*, mark size=2.0pt, solid, forget plot] coordinates {(-1.575, -10.5063)};
    \addplot [color=color_green, only marks, mark=*, mark size=2.0pt, solid, forget plot] coordinates {(-2.2797, -9.5059)};
  \end{axis}
  
  %%}

\end{tikzpicture}
    \label{fig:cin_trajectories}
  }\hfill
  \vspace{-0.9cm}
  \hspace*{-0.8cm}
  \subfloat {
    \pgfplotsset{compat=1.9}

\definecolor{color_red}{HTML}{A30D00}
\definecolor{color_green}{rgb}{0, .522, .243}
\definecolor{color_blue}{rgb}{0, 0, .9}

\begin{tikzpicture}[font=\scriptsize, trim axis right]

    \begin{groupplot}[
        group style={
            % set how the plots should be organized
            group size=1 by 2,
            % only show ticklabels and axis labels on the bottom
            x descriptions at=edge bottom,
            % set the `vertical sep' to zero
            vertical sep=7pt
        },
    grid=major,
    legend style={font=\scriptsize, row sep = -2pt, inner sep = 1 pt, outer sep = 0pt},
    x label style={at={(axis description cs:0.5,-0.3)},anchor=north},
    % % legend pos=south east,
    % legend cell align={left},
    xmin = 0.0,
    xmax = 68.0,
    scaled y ticks = false,
    % y_min = -1.0,
    % y_max = 1.0
    ]

    \nextgroupplot[
    % name=top,
    % unit vector ratio*=1 1 1,
    width=1.03\columnwidth,
    height=0.28\columnwidth,
    grid=both,
    grid style={draw=gray!30,line width=.1pt},
    ylabel= $\ddot{\omega_y}$ $\left(\frac{\text{rad}}{\text{s}^3}\right)$,
    y label style={at={(axis description cs:-0.09,.5)},rotate=0,anchor=south},
    % legend style={font=\scriptsize},
    % legend pos=south west,
    % legend cell align={left},
    % y_min = -1.0,
    % y_max = 1.0
    % xmin=-50, xmax=50,
    % axis equal image,
    %ytick={0,1,2,3,4},
    ]
    \addplot[smooth, color=color_green, line width=1.0pt] table[x=time, y=yaw_jerk] {fig/tikz_data/jerk_yaw_time_no_cin2.txt};
    \addplot[smooth, color=color_blue, line width=1.0pt] table[x=time, y=yaw_jerk] {fig/tikz_data/jerk_yaw_time_cinematography2.txt};
    % \addlegendentry{no cinematography}
    % \addlegendentry{cinematography}

    \nextgroupplot[
    name=bottom,
    % unit vector ratio*=1 1 1,
    width=1.03\columnwidth,
    height=0.28\columnwidth,
    grid=both,
    grid style={draw=gray!30,line width=.1pt},
    xlabel= Time (s),
    ylabel= $\ddot{\omega_p}$ $\left(\frac{\text{rad}}{\text{s}^3}\right)$,
    y label style={at={(axis description cs:-0.09,.5)},rotate=0,anchor=south},
    y tick label style={/pgf/number format/fixed},
    % xmin=-50, xmax=50,
    % axis equal image,
    ytick={-0.03,0.0,0.03},
    ]
    \addplot[smooth, color=color_green, line width=1.0pt] table[x=time, y=pitch_jerk] {fig/tikz_data/jerk_pitch_time_no_cin2.txt};  
    \addplot[smooth, color=color_blue, line width=1.0pt] table[x=time, y=pitch_jerk] {fig/tikz_data/jerk_pitch_time_cinematography2.txt};  

    \end{groupplot}
  
\end{tikzpicture}
    \label{fig:cin_trajectories_pitch}
  }
  \vspace{-0.2cm}
  \caption{Trajectories for the camera carrying UAV while monitoring a worker on a transmission tower. For simplicity, only the lateral shot and the first fly-over shot are shown. We compare the trajectories generated by our method (blue) with those from a baseline approach without cinematographic costs (green). The upper image displays a top view of the UAV's and target's trajectories. The small white dots on the trajectories depict transition points sampled every \SI{5}{\second} to give a notion of the speed. The bottom image depicts the temporal evolution of the jerk of the camera angles}. 
    \label{fig:flyby_shots}
    \vspace{-0.55cm}
\end{figure}

The fly-over shots were selected to film the operation as they impose relative motion between the camera and the target. This feature is regarded as richer from a cinematographic point of view. We further demonstrate how our method is able to execute these relative movements more aesthetically than a baseline approach where the specific term to smooth variations in camera angles has been removed (i.e., $\alpha_1=0$ in Problem \ref{eq:formulation_initial}). \autoref{fig:flyby_shots} compares the trajectories for the camera carrying UAV generated with both our method and the baseline approach. The baseline approach generates straight trajectories, whereas our method results in orbital trajectories, which have been used in the cinematography literature to produce more pleasant videos. For instance, \cite{Lino2015,galvane_tg18,Bucker2020} apply the arcball principle~\cite{arcball} to create a spherical surface around the target for aesthetic camera motion. We can also see in~\autoref{fig:flyby_shots} that our method reduces the jerk of the camera angles. Note that in aerial cinematography literature, the jerk of the camera motion (third derivative of the angles) has been identified as a key aspect for shot quality~\cite{bonatti_jfr20,gebhardt_chi16}.   
We measured the root mean square of the jerk of $\varphi$ and $\xi$ along the full trajectories and obtained \SI{0.0197}{\radian \per \cubic \second} and \SI{0.0048}{\radian \per \cubic \second}, respectively, for our method; and \SI{0.0265}{\radian \per \cubic \second} and \SI{0.0075}{\radian \per \cubic \second}, respectively, for the baseline without the cinematographic cost term. 

\begin{figure}
    \centering
    \vspace{0.3cm}
    \input{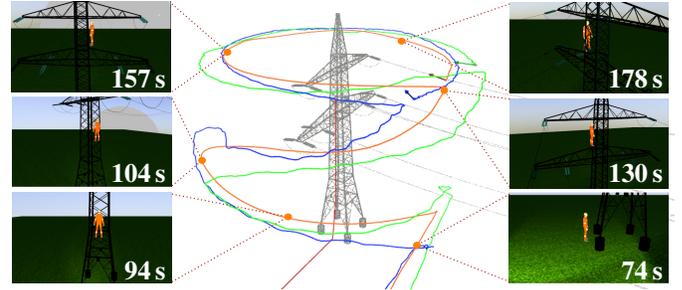}
    \vspace{-0.5cm}
    \caption{An illustration of the experiment where an operator is filmed working on a transmission tower. The trajectories of the camera carrying leader (orange), both followers carrying lights (blue and green), and the human worker (red) are shown. The obstacle map is represented by a point cloud, including the power lines and tower. The worker is tracked with a lateral shot as he walks to the tower and then with a sequence of fly-over shots while he climbs up. Several onboard images taken during the experiment are also shown.}
    \label{fig:sequence_trajectories}
\end{figure}

\begin{figure}
\vspace{0.2cm}
  \pgfplotsset{compat=1.9}

\definecolor{color_red}{HTML}{A30D00}
\definecolor{color_green}{rgb}{0, .522, .243}
\definecolor{color_blue}{rgb}{0, 0, .9}

\begin{tikzpicture}[font=\scriptsize, trim axis right]

    \begin{groupplot}[
        group style={
            % set how the plots should be organized
            group size=1 by 3,
            % only show ticklabels and axis labels on the bottom
            x descriptions at=edge bottom,
            % set the `vertical sep' to zero
            vertical sep=5pt
        },
    grid=major,
    legend style={font=\scriptsize, row sep = -2pt, inner sep = 1 pt, outer sep = 0pt},
    x label style={at={(axis description cs:0.5,-0.3)},anchor=north},
    % % legend pos=south east,
    % legend cell align={left},
    xmin = 0.0,
    xmax = 199.0,
    scaled y ticks = false
    % y_min = -1.0,
    % y_max = 1.0
    ]

    \nextgroupplot[
    % name=top,
    % unit vector ratio*=1 1 1,
    width=1.04\columnwidth,
    height=0.28\columnwidth,
    grid=both,
    grid style={draw=gray!30,line width=.1pt},
    ylabel= $d_{F}$ (m),
    y label style={at={(axis description cs:-0.09,.5)},rotate=0,anchor=south},
    % y tick label style={/pgf/number format/fixed}
    % legend style={font=\scriptsize},
    % legend pos=south west,
    % legend cell align={left},
    % y_min = -1.0,
    % y_max = 1.0
    % xmin=-50, xmax=50,
    % axis equal image,
    ytick={3.0, 4.0},
    yticklabels={3.0, 4.0}
    ]
    \addplot[smooth, color=color_green, line width=1.0pt] table[x=time, y=fov_dist_1] {fig/tikz_data/results_sim_2.txt};
    \addplot[smooth, color=color_blue, line width=1.0pt] table[x=time, y=fov_dist_2] {fig/tikz_data/results_sim_2.txt};
    % \addlegendentry{no cinematography}
    % \addlegendentry{cinematography}

    \nextgroupplot[
    % name=top,
    % unit vector ratio*=1 1 1,
    width=1.04\columnwidth,
    height=0.28\columnwidth,
    grid=both,
    grid style={draw=gray!30,line width=.1pt},
    ylabel= $\varphi_{d}$ (rad),
    y label style={at={(axis description cs:-0.09,.5)},rotate=0,anchor=south},
    y tick label style={/pgf/number format/fixed},
    legend columns = -1,
    % legend style={font=\scriptsize},
    % legend pos=south west,
    % legend cell align={left},
    % y_min = -1.0,
    % y_max = 1.0
    % xmin=-50, xmax=50,
    % axis equal image,
    ytick={0.0, 0.07, 0.14},
    ]
    \addplot[smooth, color=color_green, line width=1.0pt] table[x=time, y=heading_diff_1] {fig/tikz_data/results_sim_2.txt};
    \addplot[smooth, color=color_blue, line width=1.0pt] table[x=time, y=heading_diff_2] {fig/tikz_data/results_sim_2.txt};
    \addlegendentry{light 1}
    \addlegendentry{light 2}

    \nextgroupplot[
    name=bottom,
    % unit vector ratio*=1 1 1,
    width=1.04\columnwidth,
    height=0.28\columnwidth,
    grid=both,
    grid style={draw=gray!30,line width=.1pt},
    xlabel= Time (s),
    ylabel= $\xi_{d}$ (rad),
    y label style={at={(axis description cs:-0.09,.5)},rotate=0,anchor=south},
    % y tick label style={/pgf/number format/fixed}
    % xmin=-50, xmax=50,
    % axis equal image,
    ytick={0.0, 0.04, 0.08},
    yticklabels={0.0, 0.04, 0.08},
    ]
    \addplot[smooth, color=color_green, line width=1.0pt] table[x=time, y=pitch_diff_1] {fig/tikz_data/results_sim_2.txt};  
    \addplot[smooth, color=color_blue, line width=1.0pt] table[x=time, y=pitch_diff_2] {fig/tikz_data/results_sim_2.txt};  

    \end{groupplot}
  
\end{tikzpicture}
  \vspace{-0.4cm}
  \caption{Temporal evolution of the distance $d_F$ of UAVs carrying lights from entering the camera FoV, deviation from desired heading $\varphi_d$ and deviation from desired pitch angle of light $\xi_d$.}
  \label{fig:simulation_stats}
    \vspace{-0.55cm}

\end{figure}
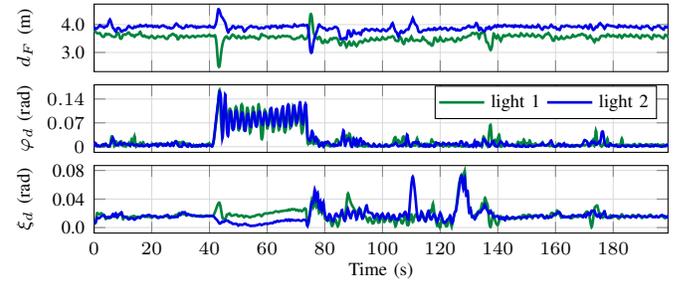

\autoref{fig:sequence_trajectories} shows the trajectories followed by the whole UAV formation throughout the experiment to film the maintenance operation. 
It can be seen that the formation is properly maintained to avoid collisions between the UAVs and the tower, and to provide required lighting of the filmed object. Moreover, none of the UAVs appear in the camera's field of view. 
The temporal evolution of the deviations from the desired orientation of each light and their distance from entering the camera FoV during this simulation are shown in~\autoref{fig:simulation_stats}.
A video of the complete simulation can be found at the site with multimedia materials. 

\subsection{Simulation - Cluttered environment}

\begin{figure}
    \centering
    \input{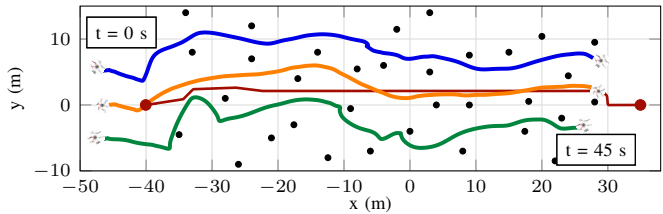}
    \vspace{-0.5cm}
    \caption{A top view of the trajectories generated in the cluttered forest scenario. The trajectories of the target (red), the leader (orange), and both followers (blue and green) are shown. The black dots represent trees.}
    \label{fig:forest_simulation}
    \vspace{-0.55cm}

\end{figure}

The aim of this simulation was to demonstrate the performance of our method for trajectory planning in a cluttered environment while assessing its scalability with numerous obstacles. We simulated a forest-like scenario with multiple trees as obstacles. As a human target walks through the forest, the filming UAV executes a chase shot from behind while the lighting UAVs follow the leader side by side. 
\autoref{fig:forest_simulation} depicts the distribution of the obstacles around the forest and the generated trajectories for the UAVs. In this figure, it is visible that the UAVs were able to follow the human in formation and to simultaneously avoid obstacles.

Finally, we analyze the scalability of our method in terms of computational demand. Simulations were run with a 4-core Intel(R) Core(TM) i7-10510U CPU @ 1.80 GHz. Table~\ref{tab:computational_results} shows the results of our method that correspond to the total planning time for each iteration that was run on the leader UAV. As expected, most time was spent during the non-convex optimization step described in \autoref{sec:initial_path}. The results for the followers are not included because they skip this non-convex optimization and thus, consume less time. The results are similar for the two simulations, although the second scenario was significantly more cluttered.

Since the map of the environment is transformed into safe corridors made of convex polyhedrons, cluttered environments do not represent an increase in the computational demands of the trajectory optimization method. Therefore, we are able to plan the leader's trajectories at a rate of \SI{1}{\hertz} with horizon lengths of \SI{8}{\second}. This rate is adequate for real-time performance in the dynamic scenarios that we target. 
The lower computational complexity required to generate the initial trajectories of the followers allows us to plan follower's trajectories at a higher rate of \SI{2}{\hertz}, enabling faster reactions to changes of the leader's behaviour and thus a more efficient mutual collision avoidance. 

\begin{table}[]
\centering
\vspace{0.2cm}
\caption{The planning times of our method per iteration. The total average values are shown for the two experiments. The percentage of time consumed at each step is shown thereafter. ITG stands for the procedure indicated in~\autoref{sec:initial_path}, SCG for procedure described in~\autoref{sec:safe_corridor_generation} and FTO for trajectory optimization described in~\autoref{sec:sec:final_trajectory_generation}.}
\resizebox{\columnwidth}{!}{%
\begin{tabular}{lcccc}
\hline
\textbf{}                  & \multicolumn{4}{c}{\textbf{Time (s)}}                                 \\ \cline{2-5} 
                           & Total $(Avg \pm std)$ & ITG $(\%)$ & SCG $(\%)$ & FTO $(\%)$ \\ \cline{2-5} 
\multicolumn{1}{c}{Tower}  & $0.70923  \pm 0.10557$  & $70.9982$       & $11.81564$       & $17.18615$       \\
\multicolumn{1}{c}{Forest} & $0.71274 \pm 0.05792$   & $72.41338$       & $8.77989$       & $18.80673$       \\ \hline
\end{tabular}%
}

\label{tab:computational_results}
\end{table}

\subsection{Real world experiment}

\begin{figure}[H]
    \centering
    \input{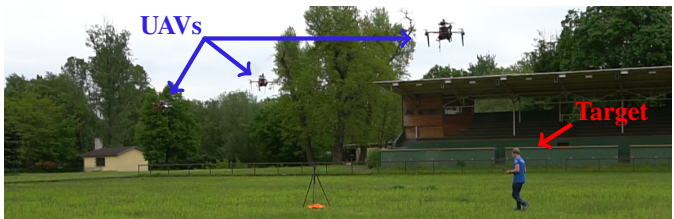}
    \vspace{-0.1cm}
    \caption{A snapshot of a real experiment in an outdoor scenario. UAVs were localized using RTK GPS. The target location was generated from a noisy ground truth, without onboard visual detection.}
    \label{fig:real_exp}
\end{figure}

In order to demonstrate our method, we performed field experiments generating trajectories for a real team of UAVs (see \autoref{fig:real_exp}). Thus, we proved the real-time performance of the proposed approach onboard. A sequence of shots was commanded to film a dynamic target in an outdoor scenario. A video of the experiment can be found at the multimedia material site.

\section{CONCLUSION}
\label{sec:conclusion}

This paper has presented a method for autonomous aerial cinematography with distributed lighting by a team of UAVs. We have proposed a novel methodology for multi-UAV trajectory planning, addressing non-linear cinematographic aspects and obstacle avoidance in separate optimization steps. We have demonstrated that the method is capable of generating smooth trajectories complying with aesthetic objectives for the filming UAV; and trajectories for the follower UAVs that allow them to keep a formation lighting the target properly and staying out of the camera FoV. Besides, our results indicate that we can plan trajectories in a distributed and online manner, and that the method is suitable for obstacle avoidance even in cluttered environments. 
As future work, we plan to address occlusions caused by obstacles within the camera FoV. Our idea is to compute the regions where these occlusions would take place and include them in the representation of the occupied space.

%%}

%%{ ACKNOWLEDGMENT

%\section*{MULTIMEDIA MATERIALS}

%The multimedia materials are available at~\url{http://mrs.felk.cvut.cz/papers/aerial-filming}.

%%}

%%{ ACKNOWLEDGMENT

% \section*{ACKNOWLEDGMENT}
%   The authors would like to thank Tom\'{a}\v{s} B\'{a}\v{c}a and Daniel He\v{r}t for they help with realization of field experiments. 
%%}

%%{ THE BIBLIOGRAPHY
\bibliographystyle{IEEEtran}
\bibliography{bibliography.bib}

% Generated by IEEEtran.bst, version: 1.14 (2015/08/26)
\begin{thebibliography}{10}
\providecommand{\url}[1]{#1}
\csname url@samestyle\endcsname
\providecommand{\newblock}{\relax}
\providecommand{\bibinfo}[2]{#2}
\providecommand{\BIBentrySTDinterwordspacing}{\spaceskip=0pt\relax}
\providecommand{\BIBentryALTinterwordstretchfactor}{4}
\providecommand{\BIBentryALTinterwordspacing}{\spaceskip=\fontdimen2\font plus
\BIBentryALTinterwordstretchfactor\fontdimen3\font minus
  \fontdimen4\font\relax}
\providecommand{\BIBforeignlanguage}[2]{{%
\expandafter\ifx\csname l@#1\endcsname\relax
\typeout{** WARNING: IEEEtran.bst: No hyphenation pattern has been}%
\typeout{** loaded for the language `#1'. Using the pattern for}%
\typeout{** the default language instead.}%
\else
\language=\csname l@#1\endcsname
\fi
#2}}
\providecommand{\BIBdecl}{\relax}
\BIBdecl

\bibitem{mademlis_tb19}
I.~{Mademlis} \emph{et~al.}, ``High-level multiple-{UAV} cinematography tools
  for covering outdoor events,'' \emph{IEEE Trans. on Broadcasting}, vol.~65,
  no.~3, pp. 627--635, 2019.

\bibitem{sabirova_20}
A.~{Sabirova} \emph{et~al.}, ``Drone cinematography system design and new
  guideline model for scene objects interaction,'' in \emph{2020 Int. Conf.
  Nonlinearity, Information and Robotics}, 2020.

\bibitem{Jeon_20}
B.~F. Jeon \emph{et~al.}, ``{Detection-Aware Trajectory Generation for a Drone
  Cinematographer},'' \emph{ArXiv e-prints}.

\bibitem{caraballo_iros20}
L.-E. Caraballo \emph{et~al.}, ``Autonomous planning for multiple aerial
  cinematographers,'' in \emph{IEEE/RSJ IROS}, 2020.

\bibitem{moreno2020jibcrane}
P.~Moreno \emph{et~al.}, ``Aerial multi-camera robotic jib crane,'' \emph{IEEE
  RA-L}, vol.~6, no.~2, pp. 4103--4108, 2021.

\bibitem{mavic}
\BIBentryALTinterwordspacing
DJI, ``Mavic pro 2,'' 2018. [Online]. Available: \url{www.dji.com/es/mavic}
\BIBentrySTDinterwordspacing

\bibitem{skydio}
\BIBentryALTinterwordspacing
Skydio, ``Skydio 2,'' 2019. [Online]. Available: \url{www.skydio.com}
\BIBentrySTDinterwordspacing

\bibitem{Bonatti2019}
R.~{Bonatti} \emph{et~al.}, ``Towards a robust aerial cinematography platform:
  Localizing and tracking moving targets in unstructured environments,'' in
  \emph{IEEE/RSJ IROS}, 2019.

\bibitem{bonatti_jfr20}
R.~Bonatti \emph{et~al.}, ``Autonomous aerial cinematography in unstructured
  environments with learned artistic decision-making,'' \emph{JFR}, vol.~37,
  no.~4, pp. 606--641, 2020.

\bibitem{naegeli_ral17}
T.~{Nägeli} \emph{et~al.}, ``Real-time motion planning for aerial videography
  with dynamic obstacle avoidance and viewpoint optimization,'' \emph{IEEE
  RA-L}, vol.~2, no.~3, pp. 1696--1703, 2017.

\bibitem{passalis_18}
N.~{Passalis} \emph{et~al.}, ``Deep reinforcement learning for frontal view
  person shooting using drones,'' in \emph{IEEE EAIS}, 2018.

\bibitem{dang_20}
Y.~{Dang} \emph{et~al.}, ``Imitation learning-based algorithm for drone
  cinematography system,'' \emph{IEEE Trans. Cogn. Devel. Syst.}, pp. 1--1,
  2020.

\bibitem{huang_icra19}
C.~Huang \emph{et~al.}, ``Learning to capture a film-look video with a camera
  drone,'' in \emph{IEEE ICRA}, 2019, pp. 1871--1877.

\bibitem{Gschwindt2019}
M.~Gschwindt \emph{et~al.}, ``Can a robot become a movie director? {L}earning
  artistic principles for aerial cinematography,'' in \emph{IEEE/RSJ IROS},
  2019.

\bibitem{naegeli_tg17}
T.~N{\"{a}}geli \emph{et~al.}, ``{Real-time planning for automated multi-view
  drone cinematography},'' \emph{ACM Trans. Graph.}, vol.~36, no.~4, pp. 1--10,
  2017.

\bibitem{galvane_tg18}
Q.~Galvane \emph{et~al.}, ``Directing cinematographic drones,'' \emph{ACM
  Trans. Graph.}, vol.~37, no.~3, pp. 1--18, 2018.

\bibitem{Bucker2020}
A.~Bucker \emph{et~al.}, ``{Do You See What I See? Coordinating multiple aerial
  cameras for robot cinematography},'' \emph{arXiv}, 2020.

\bibitem{alcantara_access20}
A.~Alcántara \emph{et~al.}, ``Autonomous execution of cinematographic shots
  with multiple drones,'' \emph{{IEEE Access}}, pp. 201\,300--201\,316, 2020.

\bibitem{alcantara2020optimal}
------, ``Optimal trajectory planning for cinematography with multiple unmanned
  aerial vehicles,'' \emph{RAS}, vol. 140, p. 103778, 2021.

\bibitem{hall2015understanding}
B.~Hall, \emph{Understanding cinematography}.\hskip 1em plus 0.5em minus
  0.4em\relax Crowood, 2015.

\bibitem{petracek2020ral}
P.~{Petr\'{a}\v{c}ek} \emph{et~al.}, ``{Dronument: System for Reliable
  Deployment of Micro Aerial Vehicles in Dark Areas of Large Historical
  Monuments},'' \emph{IEEE RA-L}, vol.~5, no.~2, pp. 2078--2085, 2020.

\bibitem{petrlik2020ral}
M.~{Petrl\'{i}k} \emph{et~al.}, ``A robust {UAV} system for operations in a
  constrained environment,'' \emph{IEEE RA-L}, vol.~5, no.~2, pp. 2169--2176,
  2020.

\bibitem{saska2017etfa}
M.~Saska \emph{et~al.}, ``Documentation of dark areas of large historical
  buildings by a formation of unmanned aerial vehicles using model predictive
  control,'' in \emph{IEEE ETFA}, 2017.

\bibitem{kratky2020ral}
V.~{Kr\'{a}tk\'{y}} \emph{et~al.}, ``Autonomous reflectance transformation
  imaging by a team of unmanned aerial vehicles,'' \emph{IEEE RA-L}, vol.~5,
  no.~2, pp. 2302--2309, 2020.

\bibitem{chen_icra16}
{J. Chen} \emph{et~al.}, ``Online generation of collision-free trajectories for
  quadrotor flight in unknown cluttered environments,'' in \emph{IEEE ICRA},
  2016.

\bibitem{Mohta2018}
K.~Mohta \emph{et~al.}, ``{Fast, autonomous flight in GPS-denied and cluttered
  environments},'' \emph{JFR}, vol.~35, no.~1, pp. 101--120, 2018.

\bibitem{Liu2017}
S.~Liu \emph{et~al.}, ``{Planning dynamically feasible trajectories for
  quadrotors using safe flight corridors in 3-D complex environments},''
  \emph{IEEE RA-L}, vol.~2, no.~3, pp. 1688--1695, 2017.

\bibitem{Tordesillas2019}
J.~Tordesillas \emph{et~al.}, ``{FASTER: Fast and Safe Trajectory Planner for
  Flights in Unknown Environments},'' \emph{IEEE/RSJ IROS}, 2019.

\bibitem{baca2020mrs}
T.~B\'{a}\v{c}a \emph{et~al.}, ``The {MRS UAV} system: Pushing the frontiers of
  reproducible research, real-world deployment, and education with autonomous
  unmanned aerial vehicles,'' \emph{JINT}, vol.~26, 2021.

\bibitem{jpsHarabor2011}
D.~Harabor \emph{et~al.}, ``Online graph pruning for pathfinding on grid
  maps.'' in \emph{AAAI Conf. on Artificial Intelligence}, vol.~25, no.~1,
  2011.

\bibitem{harabor2014}
------, ``Improving jump point search,'' in \emph{Int. Conf. on Automated
  Planning and Scheduling}, vol.~24, no.~1, 2014.

\bibitem{Houska2011a}
B.~Houska \emph{et~al.}, ``{ACADO} {T}oolkit -- {A}n {O}pen {S}ource
  {F}ramework for {A}utomatic {C}ontrol and {D}ynamic {O}ptimization,''
  \emph{Optimal Control Applications and Methods}, vol.~32, no.~3, pp.
  298--312, 2011.

\bibitem{Lino2015}
C.~Lino \emph{et~al.}, ``{Intuitive and Efficient Camera Control with the Toric
  Space},'' \emph{ACM Trans. Graph.}, 2015.

\bibitem{arcball}
K.~Shoemake, ``Arcball: a user interface for specifying three-dimensional
  orientation using a mouse,'' in \emph{Proceedings of Graphics Interface},
  1992.

\bibitem{gebhardt_chi16}
C.~Gebhardt \emph{et~al.}, ``{Airways: Optimization-Based Planning of Quadrotor
  Trajectories according to High-Level User Goals},'' in \emph{Proceedings of
  the Conf. on Human Factors in Computing Systems}, 2016.

\end{thebibliography}

%%}

\end{document}